\documentclass[letterpaper, 10 pt, conference]{ieeeconf} 
\usepackage{amsmath,amsfonts}
\usepackage{algorithmic}
\usepackage{algorithm}
\usepackage{array}
\usepackage{textcomp}
\usepackage{stfloats}
\usepackage{url}
\usepackage{verbatim}
\usepackage{graphicx}
\usepackage{cite}
\usepackage{nomencl}
\usepackage{ntheorem}   
\usepackage{lipsum}     
\usepackage{graphicx}

\usepackage{xcolor}
\usepackage{hyperref}

\hypersetup{  
   colorlinks=true,
	linkcolor=blue,
	filecolor=green, 
	urlcolor=black,
	citecolor=orange
}

\usepackage{bm}

\DeclareRobustCommand{\uvec}[1]{{%
		\ifcsname uvec#1\endcsname
		\csname uvec#1\endcsname
		\else
		\bm{\mathbf{#1}}%
		\fi
}}

\usepackage{amssymb}

\IEEEoverridecommandlockouts                              

\usepackage{graphicx} 

\title{\LARGE \bf
MonoRollBot: 3-DOF Spherical Robot with Underactuated Single Compliant Actuator Design
}

\author{Zhiwei Liu and Seyed Amir Tafrishi 
\thanks{Zhiwei Liu and Seyed Amir Tafrishi are with Geometric Mechanics and Mechatronics in Robotics (gm$^2$R) Lab, the School of Engineering, Cardiff University, Cardiff, CF24 3AA, United Kingdom.
        {\tt\small \{liuz113, tafrishisa\}@cardiff.ac.uk}}
}
\renewcommand{\baselinestretch}{.93}
\begin{document}

\maketitle
\thispagestyle{empty}
\pagestyle{empty}


\begin{abstract}
Spherical rolling robots have garnered significant attention in the field of mobile robotics for applications such as inspection and space exploration. Designing underactuated rolling robots poses challenges in achieving multi-directional propulsion with high degrees of freedom while utilizing a limited number of actuators. This paper presents the MonoRollBot, a novel 3-degree-of-freedom (DOF) spherical robot that utilizes an underactuated mechanism driven by only a single spring-motor system. Unlike conventional spherical robots, MonoRollBot employs a minimalist actuation approach, relying on only one motor and a passive spring to control its locomotion. The robot achieves 3-DOF motion through an innovative coupling of spring dynamics and motor control. In this work, we detail the design of the MonoRollBot and evaluate its motion capabilities through design studies. We also study its locomotion behaviours based on changes in rotating mass and stiffness properties. 
\end{abstract}

\section{Introduction}
Rolling robots offer a unique solution for tasks such as inspection and exploration, particularly in environments that demand minimal interaction between the robot’s exterior and its surroundings \cite{armour2006rolling,tafrishi2019design}. These robots can operate without relying on external actuators, instead employing various internal actuation principles. Underactuation, where the robot achieves complex motion with a minimal number of actuators, presents a fascinating challenge in designing such systems \cite{tafrishi2019design}. Achieving efficient locomotion through underactuation is especially difficult in spherical robots, requiring a deep understanding of mechatronics and the physics of motion.

Spherical robots have evolved through various propulsion principles, each offering unique advantages and challenges. One prominent method is torque-reaction propulsion, where robots utilize motor-driven wheels or mechanisms to create reactive forces that enable motion. For example, Halme et al. demonstrated a kinematic control method in 1996, utilizing a single-direction turning wheel for locomotion \cite{HalmeMotion1996} or having a cart to move the spherical shell \cite{BicchiSpherenonholonomy}.  

Another propulsion technique is mass imbalance, which involves manipulating the robot's center of mass to achieve movement. Javadi introduced a mass-imbalance-driven robot in 2002, leveraging the shifting of weights along different axes to generate driving forces \cite{AugustJavadi2002}. This concept was further exemplified by NASA's Tumbleweeds rover, designed for space exploration, which combined wind energy with mass imbalance \cite{PolarNASA}. Although effective, mass-imbalance systems can face limitations, such as reduced velocity and constrained internal volume which was resolved by having an isolated rating mass system using fluid actuation by Tafrishi et al \cite{tafrishi2019design}.

A third propulsion principle is the conservation of angular momentum, where internal mechanisms like gyroscopes generate motion through the conservation of rotational forces. The Gyrover, introduced in 1996, utilized a rotating internal gyroscope to create propulsion \cite{BrownCMUGyrosignlewheel}. This principle was further advanced with the development of the Gyrosphere robot, which combined both angular momentum and torque-reaction forces for enhanced locomotion capabilities \cite{MITSchroll2008}. Each of these propulsion methods highlights the ongoing innovation and challenges in the design of spherical robots, contributing to their diverse applications.

Underactuated robotic systems, though more complex than fully actuated ones, offer advantages like energy efficiency, simpler designs, and reduced hardware complexity \cite{he2019underactuated,liu2020survey}. These systems are particularly useful in mobile robots, where minimizing actuators lowers energy consumption and control complexity \cite{tafrishi2019design,dong2020back}. By leveraging dynamic interactions with the environment, underactuated robots achieve effective locomotion with fewer actuators, similar to biological systems like Armadillo, which use minimal actuation for adaptive rotational movements. However, achieving versatile locomotion with a single actuator, especially in rolling robots, presents significant challenges which hardly explored. Controlling multiple DOF with just one actuator demands complex mechanical design for utilise as much as possible with precise coordination of passive dynamics and interaction with the environment.
\begin{figure}[t!]
      \centering
      \includegraphics[width = 0.45\textwidth]{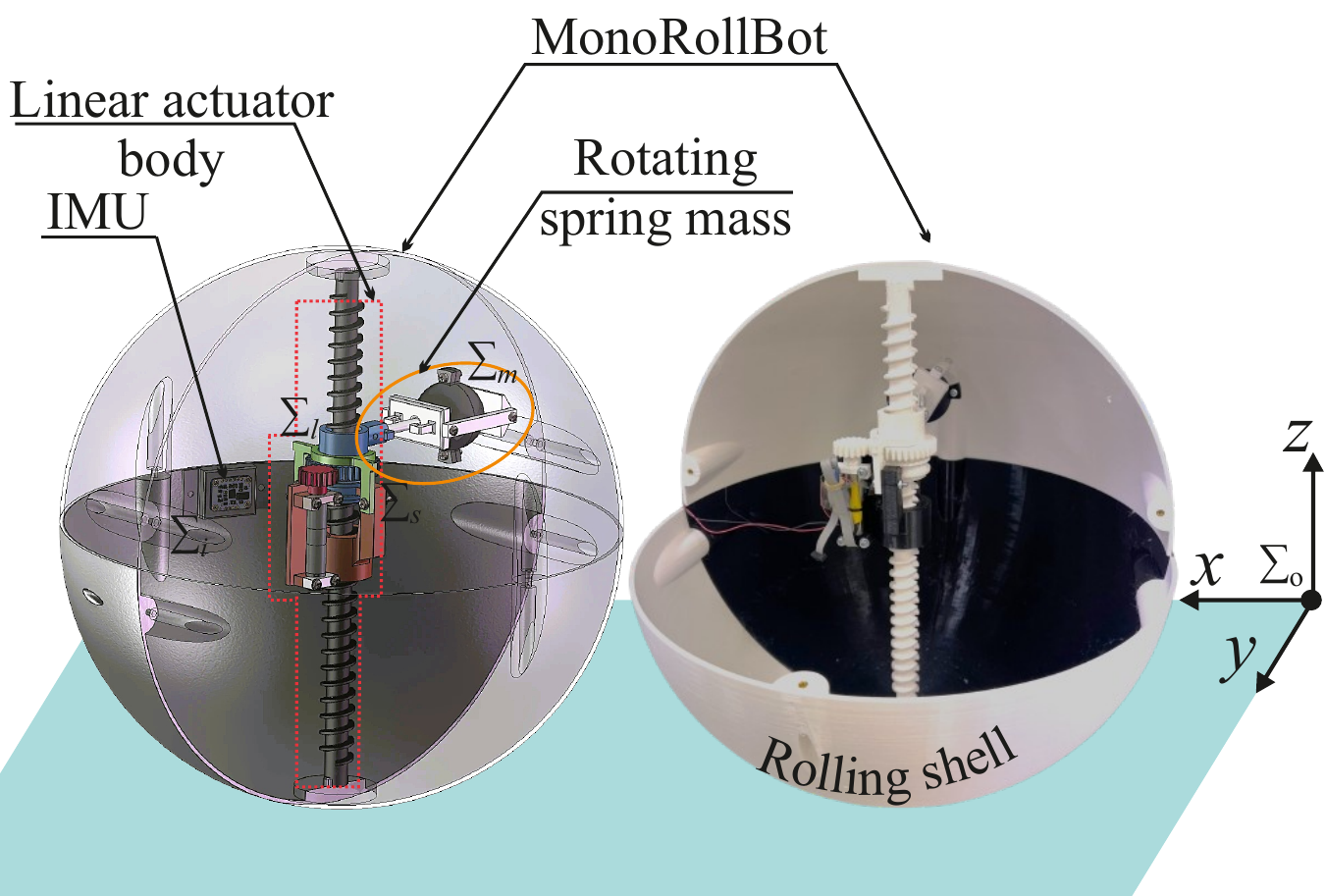}
      \caption{MonoRollBot complete robot design and details.}
      \label{fig:robot_design}
\end{figure}

The motivation for this paper is to investigate robots with multiple DoFs that operate using a single actuator and compliant mechanisms. We introduce a novel compliant underactuated robot, MonoRollBot as shown in Fig. \ref{fig:robot_design}, designed to achieve 3-DoF spherical motion, marking one of the first examples of its kind. The robot features a compliant rotating mass mechanism that facilitates effective spring-mass manipulation for movement generation, thereby reducing the need for multiple actuators. We also propose an estimation method to track the position of the rotating mass inside the rolling robot using a motor encoder and IMU data. Additionally, we conduct a detailed motion analysis focusing on two key factors: the mass of the internal rotating element and the stiffness of the spring mechanism. By systematically varying these parameters, we examine the robot's dynamic performance, providing insights into how variations in mass and stiffness influence its motion capabilities.

The paper is organized as follows: Section II discusses the design of MonoRollBot, including the mathematical modeling of the robot, and introduces estimation methods for tracking the rotating mass. Section III presents motion studies analyzing the impact of mass and stiffness variations on the robot's motion performance.


\begin{figure}[t!]
\vspace{.2 in}
      \centering
      \includegraphics[width = 0.42\textwidth]{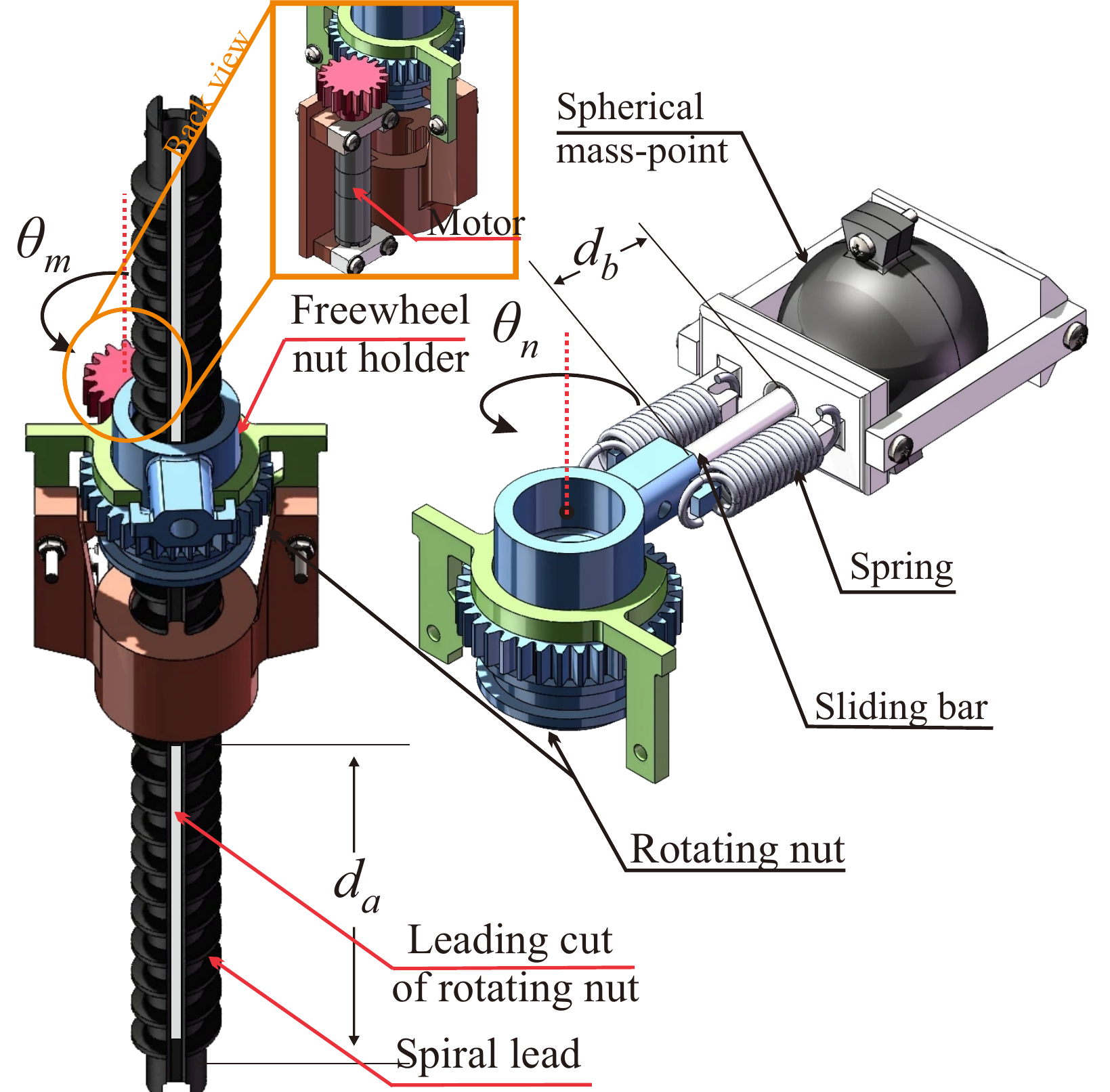}
      \caption{Actuating mechanism for 3DoF motion locomotion by MonoRollBot.}
      \label{fig:actuationmechanism}
\end{figure}
\section{Robot Design}
\subsection{Overall Structure}

The robot, referred to as MonoRollBot, is a 3-DOF spherical robot designed to achieve rolling and balancing within a spherical shell as shown in Fig. \ref{fig:robot_design}. All components of the robot, except for the motor, are fabricated using 3D printing technology. Polylactic acid (PLA) is used as the printing material due to its mechanical efficiency and sufficient strength, which are essential for maintaining the structural integrity and operation of the robot while minimizing its weight. The robot's body integrates both a linear actuator and a rotating mass system, forming the core mechanism for movement and stability on $\Sigma_l$ frame. The key design goal is to enable spherical motion with an underactuated mechanism, utilizing fewer motors (only one motor here) to achieve efficient movement of mass-point on $\Sigma_m$ frame. The robot consists of a rolling shell with the frame of $\Sigma_s$ and an internal driving structure. The external rolling shell encapsulates the internal components, providing protection and omnidirectional motion on a flat surface. The movement is characterized by a global coordinate system defined by $\Sigma_o$. 

The aim is to rotate a symmetric mass in multiple directions to create 3DoF capabilities for MonoRollBot. The robot uses a combination of a linear actuator and a rotating spring-mass system to generate movement. Table \ref{tab:ParameterRo} presents all the mechanical parts and their corresponding parametric variables. The parameters in Table \ref{tab:ParameterRo} were determined through theoretical analysis, experimental validation, and design constraints considering single DC motor. Key considerations include optimizing the spiral screw pitch $P$ for balanced torque $\tau_m$ and linear displacement and selecting the motor's nominal current to meet load and power requirements. These values were iteratively refined via simulations and prototype testing to ensure optimal performance for moving mass along the sliding bar. This design reduces the number of active motors, achieving actuation through a single spring-motor combination. The motor, positioned at the central axis of the robot, drives the rotating nut through a gear system. This nut, in turn, interacts with the spiral lead to generate both linear and rotational motion as shown in Fig. \ref{fig:actuationmechanism}. The spherical mass, mounted on the rotating nut, does not spin at the same speed as the spiral lead; instead, its motion by $\theta_n$ angle is influenced by the spring dynamics and the resultant motor torque through gears. For the linear actuator body, the linear actuator is arranged along the central vertical axis of the robot, driving vertical motion and assisting in internal mass transfer as shown in Fig. \ref{fig:actuationmechanism}. The positioning of the actuator allows the centre of mass to be changed, thus affecting the rolling direction of the spherical shell. In addition, the rotating spring mass is connected to a spring-loaded mechanism that stores potential energy and releases it to assist in motion. The mass rotates around the central axis of the robot, affecting its orientation and balance, and is therefore essential for generating rotational motion.

\begin{table}[t!]
	\vspace{2mm}
	\caption{Robot parameters and properties.}
	\label{tab:ParameterRo}
        \centering
	\begin{tabular}{ccc}
		\hline
Description & Value & Parameters \\ \hline
Radius of the sphere & 0.17 m & $R$ \\ 
Mass of the sliding bar & 0.028 kg & $m_{sb}$ \\ 
Mass of the rolling shell & 1 kg & $m_s$ \\ 
Mass of the rotating nut & 0.02 kg & $m_{rn}$ \\ 
Mass of the spiral screw & 0.085 kg & $m_{sl}$ \\ 
Pitch of the spiral screw & 10 mm & $P$ \\ 
Lead of the spiral screw  & 20 mm & $l$ \\ 
Number of teeth on the rotating nut & 34 &$\eta_n$\\ 
Number of teeth on the gear of motor & 17 & $\eta_m$\\ 
Length of the spiral lead & 0.319 m & $L_{sl}$ \\ 
Length of the sliding bar & 0.11 m & $L_{sb}$ \\ 
Motor reduction ratio & 62:1 &  \\ 
Motor  nominal voltage & 6 V & \\ 
Motor nominal current & 0.3 A & \\ 
Motor torque distance from the mass &   & $\tau_m$ \\ 
Total stiffness between mass-point and link &  & $k_s$ \\
The displacing distance linear mass &  & $d_b$ \\ 
Rotating nut angle &  & $\theta_n$ \\ 
Motor angle &  & $\theta_m$ \\ 
        \hline
	\end{tabular}
\end{table}
The robot’s internal driving system consists of a spiral lead mechanism, a rotating nut, and a freewheel nut holder, which together control the spherical mass’s position across multiple locations, as shown in Fig. \ref{fig:actuationmechanism}. The redesigned spiral lead mechanism enables linear motion along the robot’s vertical axis (represented by $d_a$ as one degree of freedom). Coupled with the rotating nut, this mechanism generates rotational motion ($\theta_n$ as a second degree of freedom) and simultaneously translates linear displacement. The freewheel nut holder prevents backward motion, allowing for precise linear actuation. Additionally, the freewheel nut holder and the cylindrical motor base are designed with small rectangular buckles that align with grooves on the rotating nut, supporting the rotating nut and ensuring accurate threading between the nut and the spiral screw. A rotating mass, connected to the rotating nut by a tension spring (providing $d_b$ as a third degree of linear motion), can slide along a sliding bar, adjusting key factors such as the robot’s center of gravity and rotational inertia.

\begin{figure}[t!]
      \centering
      \includegraphics[width = 0.5\textwidth]{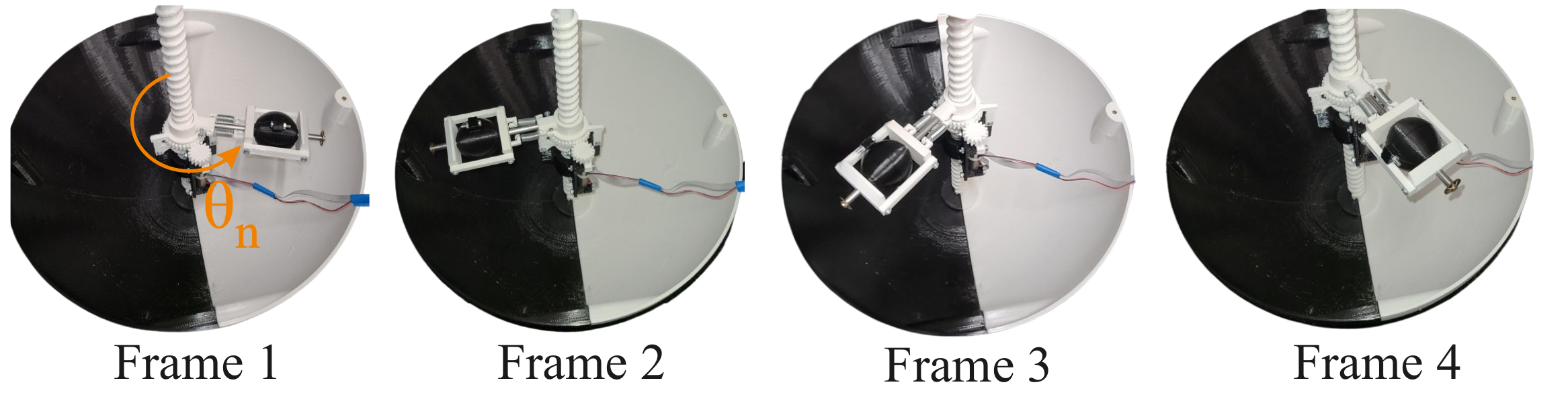}
      \caption{Actuator operation captured in snapshots where the mass rotates from the bottom of the screw lead to the midpoint of the sphere.}
      \label{fig:Snapshotmotion}
\end{figure}

One of the innovative aspects of the robot’s design is the interaction between the rotating nut and the spring-mounted mass (example rotation case shown in Fig. \ref{fig:Snapshotmotion}). As the nut rotates, rotational momentum compresses or stretches the spring, thereby storing mechanical energy. This stored energy can be released to facilitate rapid adjustments in the robot’s center of mass, enabling dynamic movements such as quick turns and rapid accelerations. The sliding bar, which connects to the rotating nut, ensures smooth motion and maintains a reliable mechanical link between the actuator and the mass point. This design allows direct control over the mass-point's motion inside the sphere with ($d_a,\theta_n$), while also indirectly adjusting the radial distance of the mass $d_b$ as momentum changes. Consequently, both the velocity and the nature of gravitational force interactions vary the radial distance through spring compression which we will carefully study in next incoming section. This design makes the robot highly responsive to the physics of locomotion, distinguishing it as unique among its kind.



\subsection{Robot Kinematics and Driving Mechanism Model}
We describe the relative kinematics and mechanism formulation, which will help identify the robot's mechanical internal actuator motion capabilities and properties. Also, we propose an estimation method for determining the position of the rotating mass in the sphere.

The MonoRollBot is equipped with an IMU attached to its shell at $\Sigma_i$, allowing calculation of the Euler angles ($\gamma, \beta, \alpha$) of the rotating body with respect to the inertial frame $\Sigma_o$ as follows:
\begin{align}
 & \uvec{R}(\gamma,\beta,\alpha)= \uvec{R}_z(\gamma) \uvec{R}_y(\beta) \uvec{R}_x(\alpha) \nonumber\\ 
 & = \begin{bmatrix}
C_\alpha C_\beta & C_\alpha S_\beta S_\gamma - S_\alpha C_\gamma & C_\alpha S_\beta C_\gamma + S_\alpha S_\gamma \\
S_\alpha C_\beta & S_\alpha S_\beta S_\gamma + C_\alpha C_\gamma & S_\alpha S_\beta C_\gamma - C_\alpha S_\gamma \\
-S_\beta & C_\beta S_\gamma & C_\beta C_\gamma 
\end{bmatrix},
\end{align}
where $C$ and $S$ are the cosine functions. Next, to obtain the amount of locomotion that happens by the rolling sphere on the plane $(x,y)$, we used linear kinematics to calculate the sphere contact location
\begin{equation}
\uvec{D} = \begin{bmatrix} x \\ y \\ 0 \end{bmatrix} = \begin{bmatrix} R (\beta-\beta_0) \\ -R (\alpha -\alpha_0)\\ 0 \end{bmatrix},
\label{Eq:motiondistance}
\end{equation}
where $R$ is the sphere radius and $(\beta_0,\alpha_0)$ are the initial angular rotation. 

Next, we need to estimate the location of the rotating mass within the sphere, as direct sensory feedback is unavailable and the rotation speed is too fast for inertial sensors to track accurately. To address this, we utilize motor encoder data and motion dynamics to determine the position of the rotating spiral nut gear. Initially, the motor torque is directly related to the springs connected to the spherical mass points including other connecting materials $m_b=m+m_{sb}$, with the following motion equation formulated for discrete computation \cite{tafrishi2022psm} by
\begin{align}
& \left[m_b d_b^2(k) \right] \ddot{\theta}_{\Delta m}+r_m\left[k_s \left(d_{b}(k)-d_{s,0}\right) \right]-m_b d_b(k) \mu_g  \nonumber\\
   &\triangleq \left[m_b d_b^2(k) \right] \frac{\dot{\theta}_m(k)-\dot{\theta}_m(k-1)}{\Delta T}+r_m\left[k_s \left(d_{b}(k)-d_{s,0}\right) \right]\nonumber\\
   &-m_b g d_b(k) \cos \alpha = \tau_m ,
   \label{Eq:MomentumTorquein}
\end{align}
where $r_m$ is the true motor torque distance from the mass link, $k_s$ is the total stiffness between mass-point and link, and ($d_b(t)$,$d_{b,0}$) are the current and initial position of the displacing linear mass. By finding the Eq. (\ref{Eq:MomentumTorquein}) for the $d_b(k)$, we will have solution of 2nd order equation, assuming always $d_b>0$, as follows
\begin{align}
    &d_b(k) = \frac{1}{{2 m_b \ddot{\theta}_{\Delta m}}}\Big[(-m_b \mu_g + k_s r_m) + \big[(m_b \mu_g)^2 \nonumber\\
    &- 2  m_b \mu_g  k_s  r_m + (k_s  r_m)^2 \nonumber\\ & +4   m_b \ddot{\theta}_{\Delta m}  k_s  r_m  d_{s,0} + 4 m_b \ddot{\theta}_{\Delta m}  \tau_m \big]^{\frac{1}{2}}\Big].
    \label{Eq:MomentumTorque}
\end{align}
\begin{figure*}[t!]
      \centering
      \includegraphics[width = 4.5 in, height = 2.7 in]{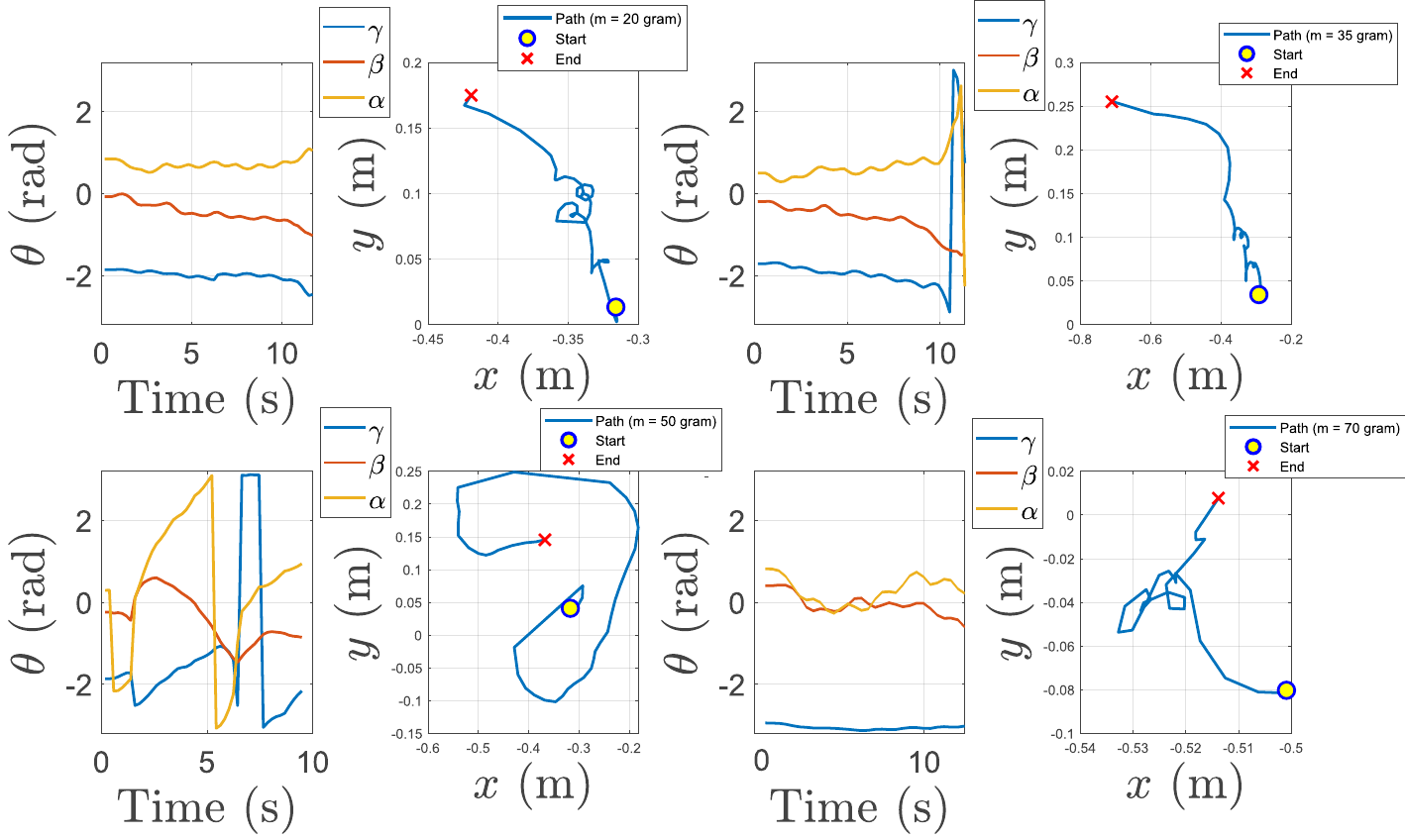}
      \includegraphics[width = .34\textwidth]{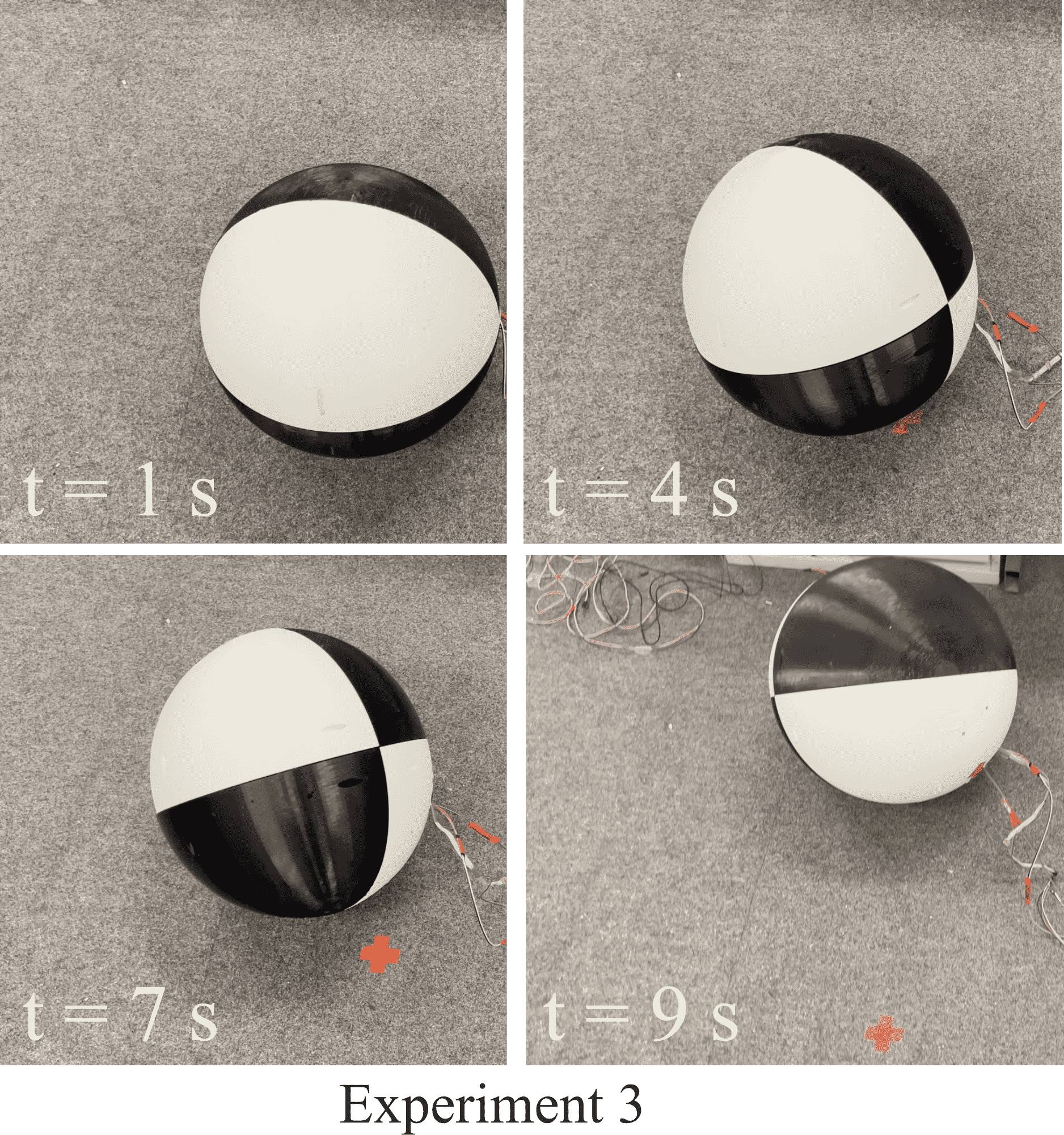}
      \caption{Actuating mechanism for 3DoF motion locomotion by MonoRollBot. (Actuated with minimum voltage to move the body mass)}
      \label{fig:2DPlotMassdifferent1}
\end{figure*}
It is important to note that in this solution, the approximated angular acceleration is captured by the motor speed from the encoder, denoted as $\ddot{\theta}_{\Delta m}$. The gravitational component $\mu_g(\alpha)$ is related to measurements from the IMU and the motor encoder. To calculate the gravitation direction angle $\alpha$, because the slider avoids the gravitational force in the axial direction only gravity force is applicable if the sliding rod is angled, this means we have to use two vectors for a linear actuator frame $\Sigma_l$ and a moving point of sphere along the same actuator, which considered as 
\begin{equation}
    \uvec{T}_d=\uvec{R}\left[\begin{array}{c}
0\\
0\\
d_a(k)
\end{array}\right] , \uvec{T}_s =\uvec{R}\left[\begin{array}{c}
R\sin\theta_d \;\cos \theta_n\\
R\sin \theta_d \;  \sin \theta_n \\
d_a(k)
\end{array}\right],
\label{Eq:vectors}
\end{equation}
where $\theta_d=2\sin^{-1}(d_a/2R)$. Then, the angle $\alpha$ can be calculated by the dot product of vectors in (\ref{Eq:vectors}) as
\begin{equation}
    \alpha= \cos^{-1} \sin \left(\frac{\uvec{T}_d\cdot\uvec{T}_s }{\parallel\uvec{T}_d\parallel\parallel\uvec{T}_s\parallel} \right)
\end{equation}
where a 3D trigonometric transformation is applied to the current vectors from $(\Sigma_l$-$\Sigma_s)$ frames at the contact point between the sphere and the ground. Additionally, we transform the linear displacement of the rotating link in $\Sigma_l$, determined by translating the motion of the spiral rod along the rotating motor gear as follows:
\begin{equation}
    d_a=l\frac{\theta_n(t)}{\pi},
\end{equation}
where $l$ is the lead of the spiral screw which $\theta_n$ can be found from gear transformation of motor gear $\theta_m$ as $\theta_n=(\eta_m/\eta_n)\;\theta_m$ as well as the angular velocity $\dot{\theta}_n=(\eta_m/\eta_n)\;\dot{\theta}_m$.

Additionally, to determine the total generated force \( F_b \) by the rolling MonoRollbot during locomotion, we utilize the filtered IMU acceleration \( \mathbf{a} \) with a zero-low pass filter $\mathbf{a}(t)$, expressed as 
\begin{equation}
F_n \approx m_b \parallel \mathbf{R}(t) \; \mathbf{a}(t) - \mathbf{R}(t) \; \mathbf{a}_g \parallel, 
\end{equation}
where \( \mathbf{a}_g \) represents the initial measurement at time \( t_0 \).

\section{Motion Study}
This section examines the capabilities of the designed robot from two perspectives: variations in rotating mass and changes in the stiffness of the connected mass. We also analyze factors such as overall behavior, total generated force, motor response, and additional parameters to assess the potential of MonoRollBot.The experiments in this section use predefined input signals to test the robot’s ability to maintain a stable trajectory while experiencing sudden changes in angular velocity and torque. The data collected in the experiments include motion trajectories, angular displacements, and system responses, as shown in Fig. \ref{fig:2DPlotMassdifferent1}.
\subsection{Rotating Mass Diversity}
At first, We explore how variations in the rotating mass $m$ impact the dynamic behaviour of MonoRollBot during locomotion. The robot's movement relies on the interaction between its internal rotating mass and the spherical shell. By altering the mass of the rotating body in different directions, we can observe distinct differences in its rolling motion, stability, and overall motion.

To investigate the effect of different rotating masses, we conducted experiments with a series of the mass of the internal rotating component varied between $\left[20-70\right]$ gram. The robot was placed on a flat surface with the same position for every experiment, and the internal motor was actuated to induce spherical motion with pre-defined length of motion on $d_a$ (being from initial to final length of spiral lead). Throughout the experiment, an IMU sensor fixed inside the robot measured the rotational angles in the global coordinate system $\left(\gamma,\beta,\alpha \right)$, while the motor encoder recorded the motor’s angular rotation $\left(\theta_m,\dot{\theta}_m\right)$

The motor speed was set to a fixed value during all experiments. The control process involved sending predefined speed commands to motor drivers through the communication interface. This approach ensured consistent and repeatable actuation during the experiments, eliminating variations caused by the dynamic control algorithm. During the operation of the experiment, the motors drove the helical lead mechanism at a specified speed. This simple control method allowed for a controlled study of the robot's kinematic dynamics under different mass and stiffness configurations, isolating the effects of the physical parameters from the complex control effects.

\begin{figure}[t!]
      \centering
      \includegraphics[width = .37\textwidth]{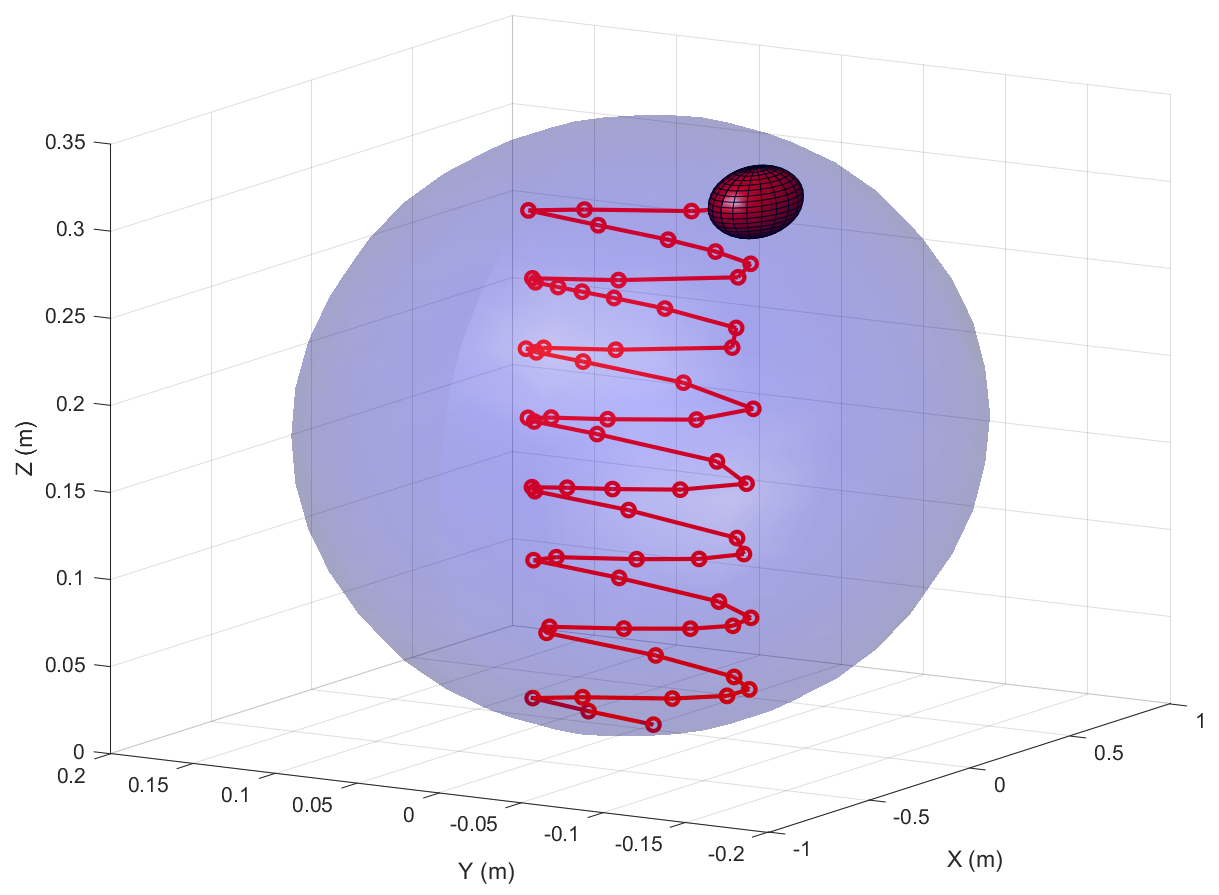}
      \caption{Experiment 3 estimation of rotating-mass location inside the sphere.}
      \label{fig:masstrajectoryexperi2m1}
\end{figure}
\begin{figure}[t!]
      \centering
      \includegraphics[width = 3.2 in, height = 1.4 in]{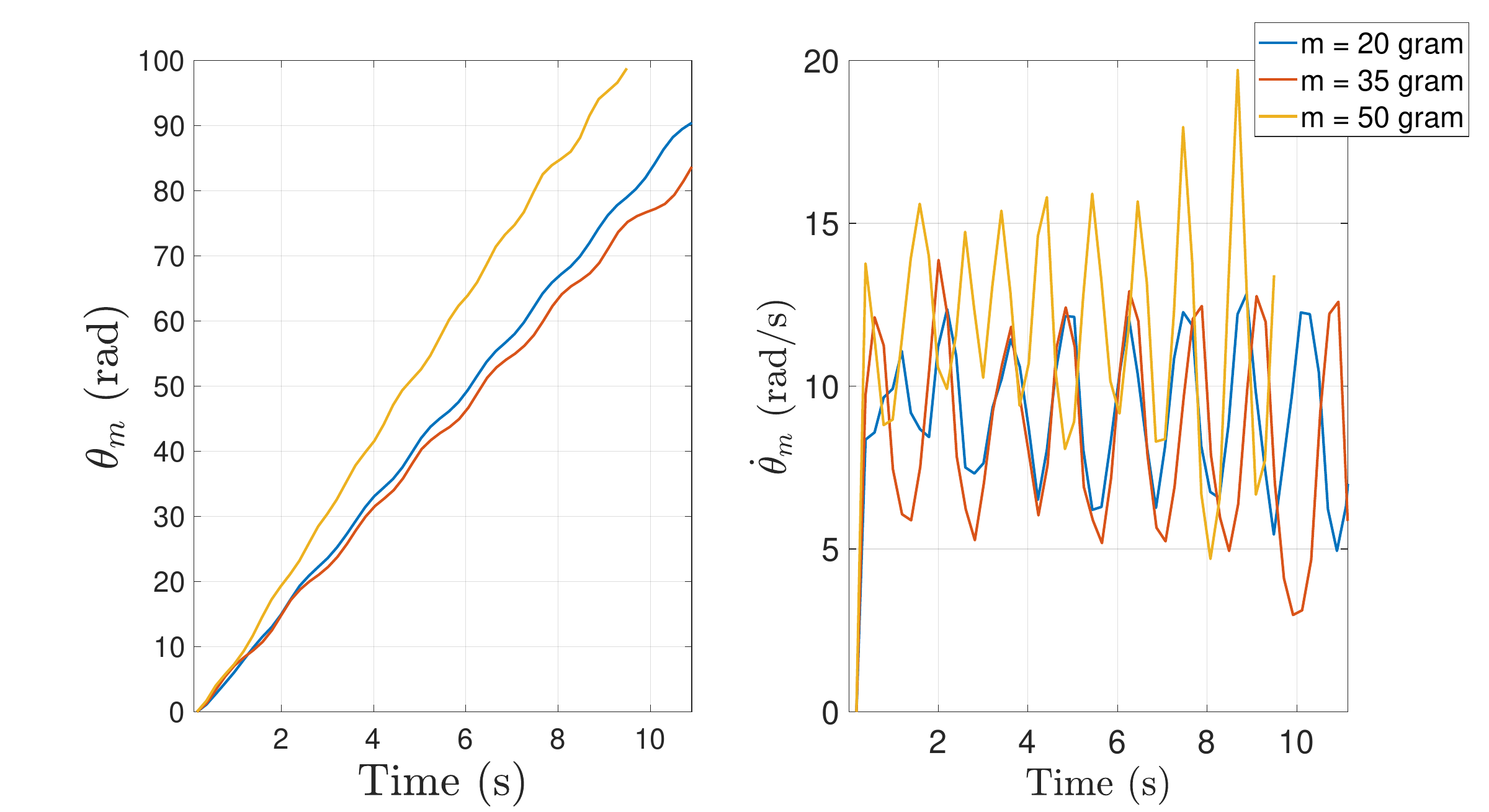}
      \caption{DC motor angular rotation and velocity for various masses.}
      \label{fig:MotorResponse1}
\end{figure}

\begin{figure}[t!]
      \centering
      \includegraphics[width = .28\textwidth]{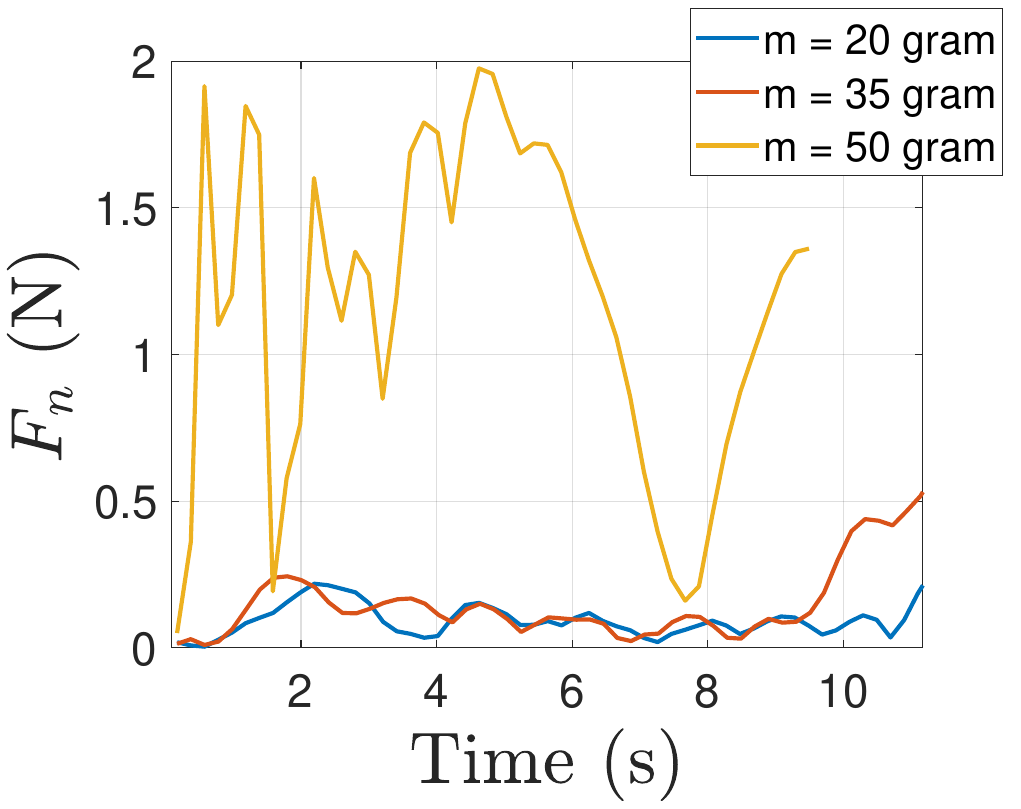}
      \caption{Total generated force created for sphere's locomotion in a different internal mass of mass-point.}
      \label{fig:Forcemomentmass1}
\end{figure}
Fig. \ref{fig:2DPlotMassdifferent1}. shows the robot's rotational angles through IMU and its path trajectory $\uvec{D}$ under different mass conditions based on Eq. (\ref{Eq:motiondistance}). For lighter masses ($20$, $35$) g, the robot's rotational response remains relatively linear motion, with smoother and less oscillatory angles $\left(\gamma,\beta,\alpha \right)$. As seen in the trajectory plots, the robot follows a more predictable and controlled path, as the lighter mass requires less effort to maintain balance and motion. The lower inertia allows the robot to adjust its orientation without causing excessive tilting or abrupt changes in direction but to propel it requires a certain conservation of angular momentum.  As the mass increases ($50, 70$) g, the rotational angles show more pronounced oscillations, especially in the $\gamma$-axis. This suggests that the larger mass creates more dynamic dependency, making the robot's motion requiring more complex dynamic-based controllers especially when mass moves the upper hemisphere as shown in Fig. \ref{fig:masstrajectoryexperi2m1} (estimated by Eq. (\ref{Eq:MomentumTorque})). In the trajectory plots, heavier masses result in larger deviations from a straight-line path, with more erratic turns/swings and fluctuations. The added inertia from the larger mass amplifies the robot's response to motor actuation, leading to quicker changes in direction or backward motion (which is minor-based on expected direction) and less smooth trajectories. In Experiment 3, the internal mass rotated at a constant velocity (Fig. \ref{fig:MotorResponse1}), while the robot’s actuator drove the system. Key snapshots highlight the interaction between the spherical mass, actuator, and spiral lead mechanism. Observed trajectory deviations and angular shifts reveal the impact of mass distribution on the robot's motion dynamics and control performance.

Fig. \ref{fig:MotorResponse1} illustrates the motor's angular position ($\theta_m$) and angular velocity ($\dot{\theta}_m$) across different mass conditions. As the rotating mass increases, the motor's angular displacement rate also increases. This effect occurs because the increased mass contributes additional inertial momentum, which allows the system to maintain motion with less motor effort. Moreover, the reaction torque produced by the added mass enables the spherical shell to respond more dynamically. For angular velocity $\dot{\theta}_m$, fluctuations become more pronounced with higher mass, likely due to inertial effects that maintain amplitude despite the increased load. This trend corresponds with the total force generated by the robot, as shown in Fig. \ref{fig:Forcemomentmass1}.

Based on these observations, we conclude that heavier masses introduce greater dynamic complexity. While they enhance the robot’s rotational responsiveness and force generation, they reduce controllability by amplifying inertia-induced deviations and oscillations. This trade-off underscores the need to select an optimal rotating mass that balances swift motion with stable control, depending on the task requirements. This approach enables choosing between actuation principles: mass-imbalance (using larger masses) or angular momentum conservation (using low to medium masses). 

\subsection{Stiffness Diversity}
Next, we explore how variations in the rotating mass impact the dynamic behavior of MonoRollBot during locomotion. The robot's movement relies on the interaction between its internal rotating mass and the spherical shell. 

To investigate the impact of spring stiffness on the robot’s movement, we conducted a series of experiments where the spring stiffness constant $k_s$ was varied with three distinct values as: $k_s$=160 N/m, $k_s$=200 N/m, and $k_s$=300 N/m. The robot was actuated under the same conditions for each stiffness value and rotating mass $m=$ 50 g, and the resulting motion was captured using the IMU sensor and the motor encoder.
 
\begin{figure}[t!]
      \centering
      \includegraphics[width = .36\textwidth]{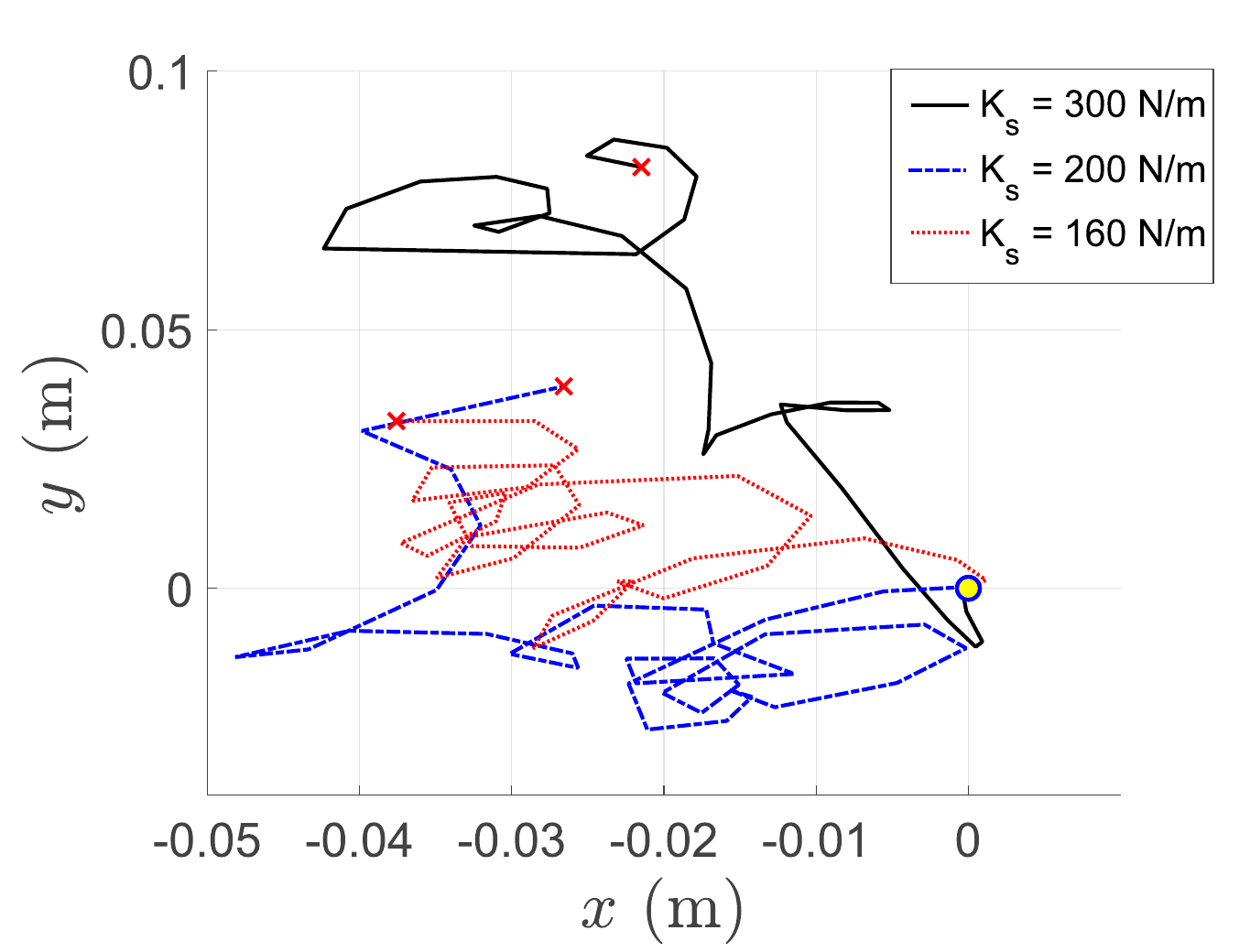}
      \caption{The trajectory of MonoRollBot on a plane under varying spring stiffness values.}
      \label{fig:2DplotStiffness1}
\end{figure}
\begin{figure}[t!]
      \centering
      \includegraphics[width = 3.2 in, height = 1.5 in]{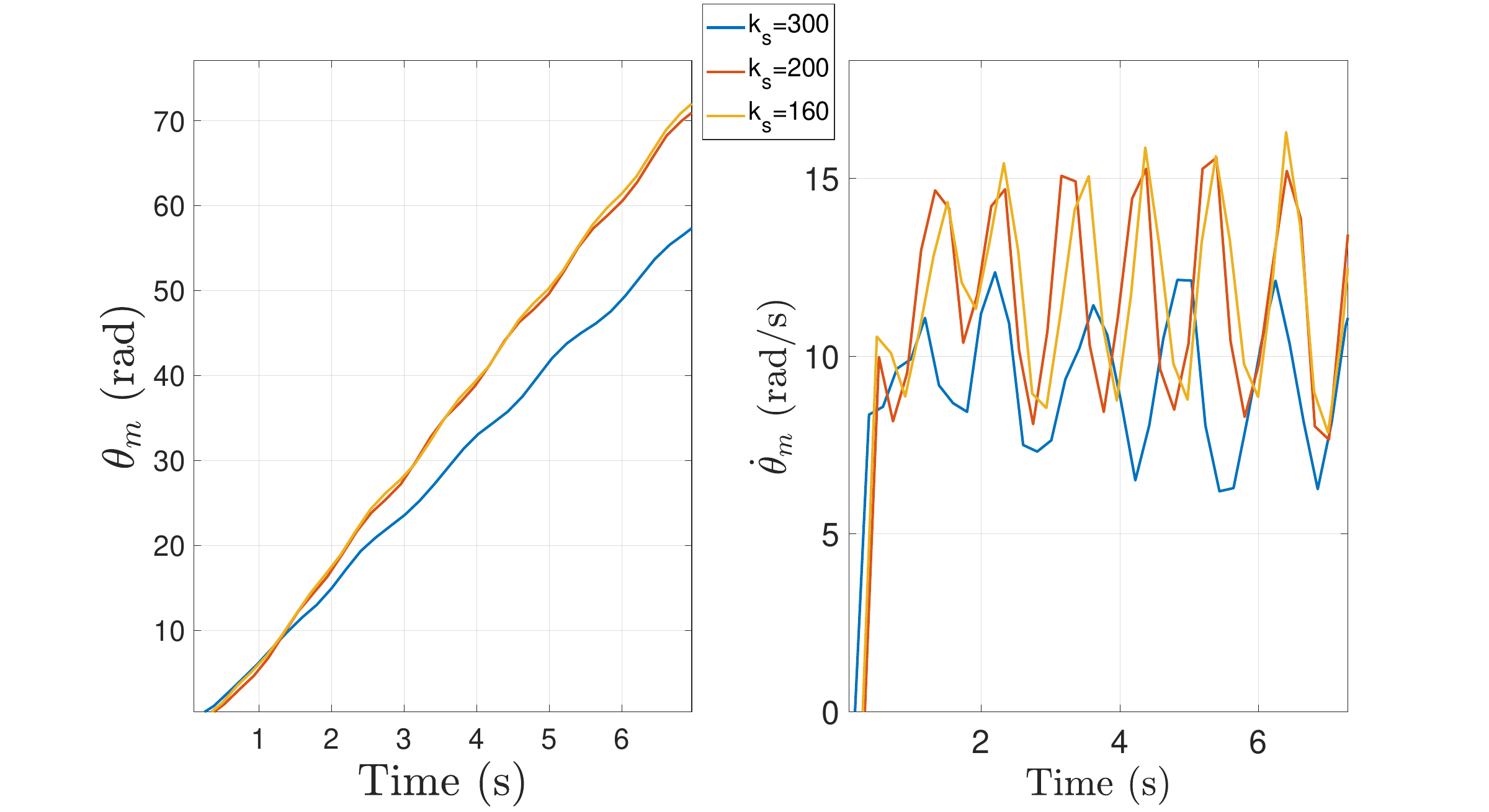}
      \caption{DC motor angular rotation and velocity for various masses.}
      \label{fig:MotorResponseStiff1}
\end{figure}
\begin{figure}[t!]
      \centering
      \includegraphics[width = .42\textwidth]{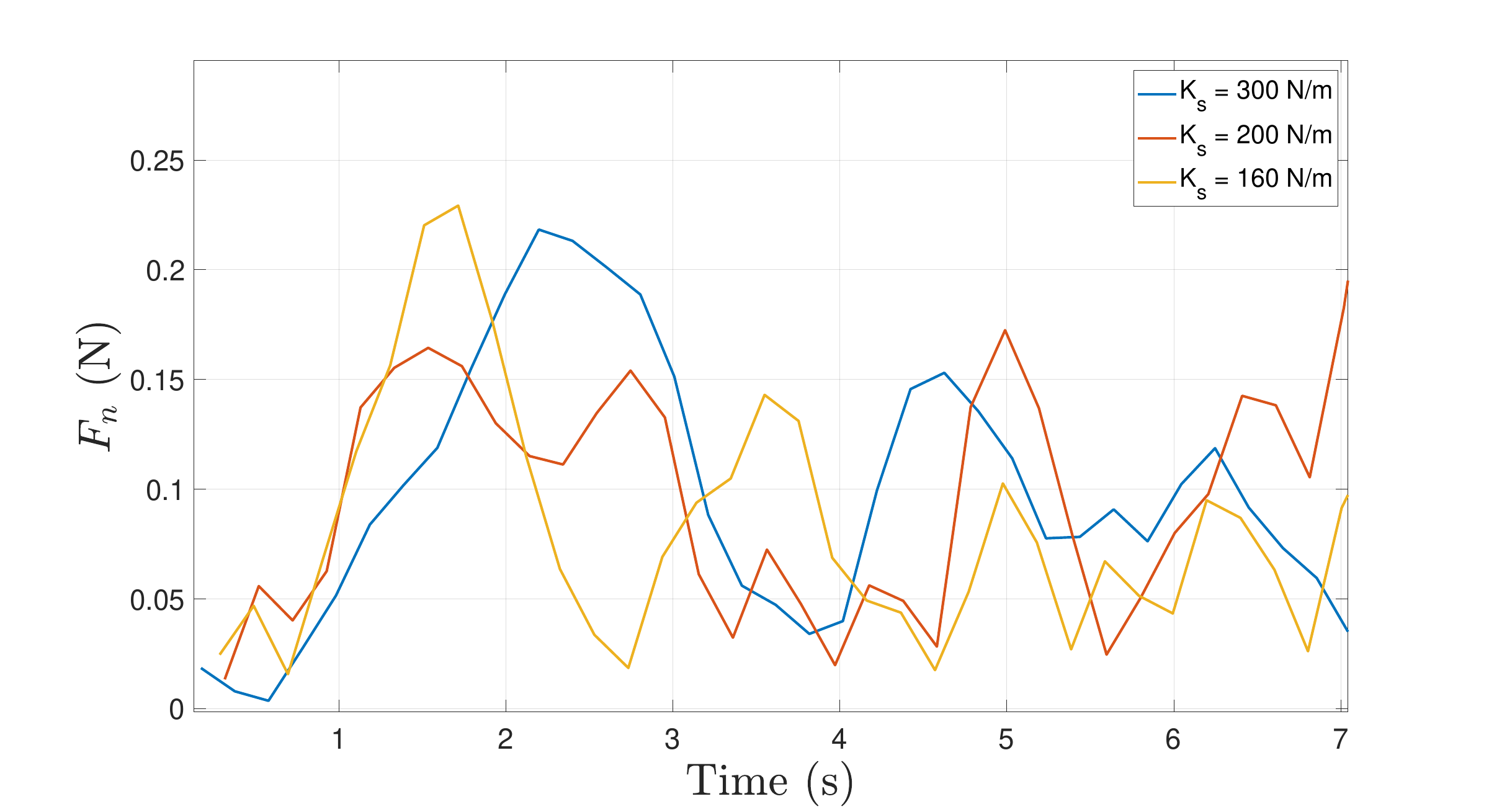}
      \caption{Total generated force created for sphere's locomotion in a different stiffness ratio of the spring.}
      \label{fig:Forcemomentstiffsness1}
\end{figure}

Fig. \ref{fig:2DplotStiffness1} shows the robot's trajectory for three spring stiffness values. With low stiffness ($k_s = 300$ N/m), the trajectory is smooth and semi-linear, with limited oscillations due to minimal spring energy storage. At medium stiffness ($k_s = 200$ N/m), the motion becomes more dynamic, allowing faster orientation changes but with increased oscillations. For high stiffness ($k_s = 160$ N/m), the robot demonstrates rapid, aggressive movements with significant trajectory deviations, where the increased energy release enhances reactivity but reduces stability. Fig. \ref{fig:MotorResponseStiff1} further shows that with higher stiffness, the motor’s angular displacement $\theta_m$ and velocity $\dot{\theta}_m$ exhibit faster motion but higher oscillation amplitudes.

\begin{figure}[t!]
      \centering
      \includegraphics[width = .3\textwidth]{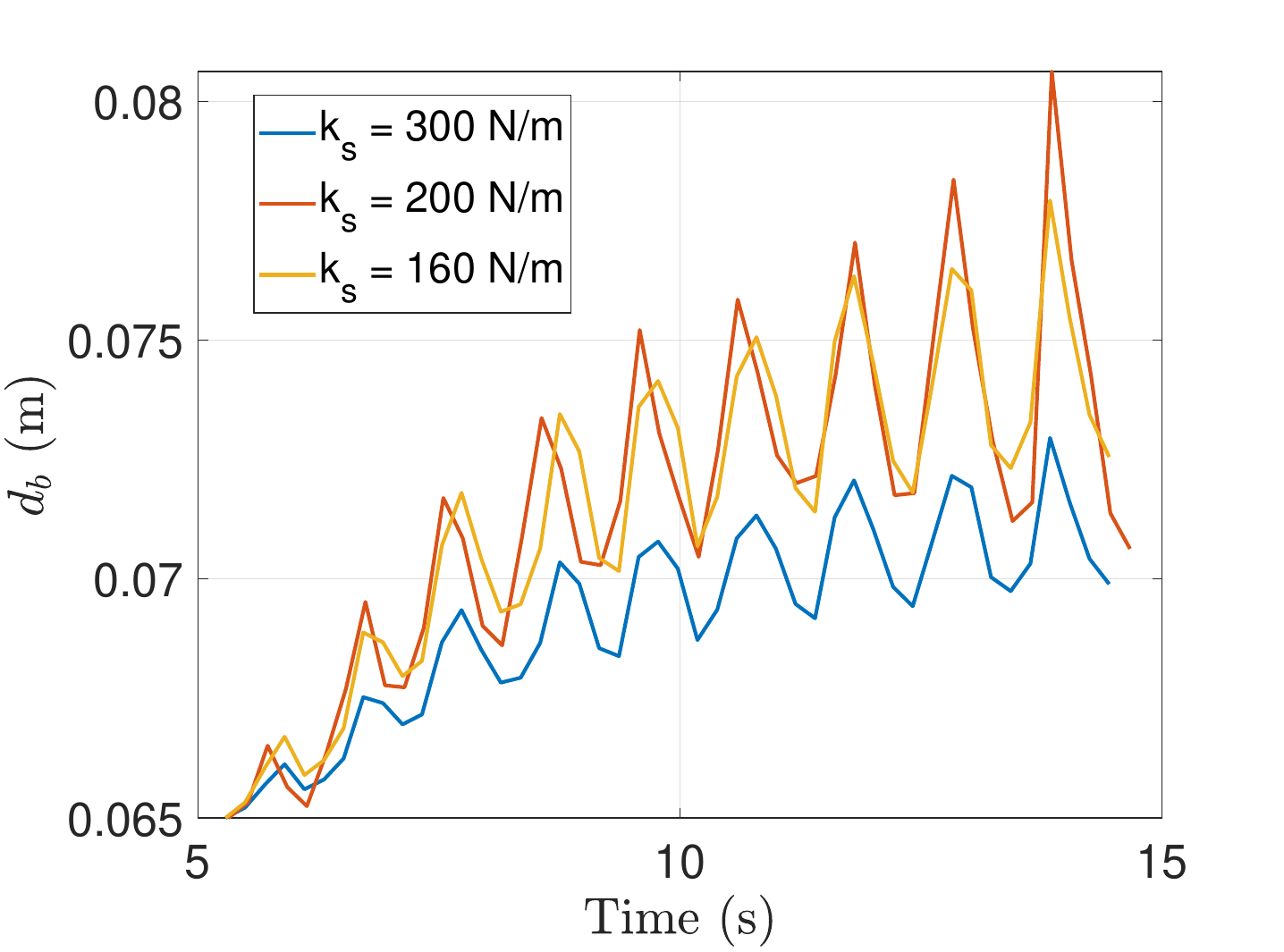}
      \caption{The variation of link with mass distance on $\Sigma_l$ frame.}
      \label{fig:StiffnessDb1}
\end{figure}

To better understand the effect of varying stiffness on the total force generated by the spherical shell, Fig. \ref{fig:Forcemomentstiffsness1} presents the data for each stiffness scenario. Higher stiffness values result in a more consistent force with lower frequency, while lower stiffness increases the frequency of force fluctuations, though the amplitude remains similar. This suggests that low stiffness may require finer motor control steps to prevent instability in the spherical shell’s motion. In contrast, high stiffness reduces instability but requires careful frequency management to maintain a continuous, strong force on the shell. Additionally, Fig. \ref{fig:StiffnessDb1} shows greater variation in $d_b$ with low stiffness, as the link mass $\Sigma_l$ shifts more dynamically, as estimated by Eq. (\ref{Eq:MomentumTorque}).

Overall, high spring stiffness enables smoother, controlled trajectories, ideal for applications requiring stability and gradual movement. In contrast, lower stiffness enhances responsiveness and force generation, allowing quicker adjustments but increasing oscillations and instability. This trade-off suggests that stiffness tuning is crucial for optimizing MonoRollBot's performance: precision tasks benefit from lower stiffness, while rapid response applications can leverage higher stiffness with careful control of oscillations. Additionally, higher stiffness supports mass-imbalance propulsion for stable motion, whereas lower stiffness may be preferable for momentum-conserving locomotion that requires quick directional changes.
 
\section{Conclusion}
This paper introduced MonoRollBot, a novel 3-DOF spherical robot driven by a single underactuated spring-motor system for efficient rolling motion. Through design studies and experiments, we analyzed how varying internal mass and spring stiffness impact the robot’s dynamics and locomotion capability. Results show that lower mass and stiffness promote smooth, stable motion, while higher values enable more dynamic but less controllable behavior, highlighting a balance between agility and stability in underactuated systems and how the actuation principle should be utilised in the rolling system. These insights provide a basis for optimizing MonoRollBot’s design for specific tasks. Future research will develop a comprehensive dynamics model using D'Alembert’s principle with nonholonomic constraints and pursue control strategies that harness MonoRollBot’s single-actuation for enhanced adaptability and multidirectional motion.





\section{Acknowledgment}
This work was supported by the Royal Society research grant under Grant \text{RGS\textbackslash R2\textbackslash 242234}.


 \bibliographystyle{ieeetr}
 \bibliography{references}

\end{document}